\documentclass[letterpaper, 10 pt, conference]{ieeeconf}  

\IEEEoverridecommandlockouts                              

\usepackage{graphicx}
\usepackage{caption}
\usepackage{subcaption}
\usepackage{pgfplots}
\usepackage[tight-spacing=true]{siunitx}
\usepackage{float}
\usepackage{amssymb,amsmath}
\usepackage{algorithmic}
\usepackage{algorithm}
\usepackage{hyperref}

\title{\LARGE \bf
Robust Gesture-Based Communication for Underwater Human-Robot Interaction in the context of Search and Rescue Diver Missions
}

\author{Arturo Gomez Chavez$^{1}$, Christian A. Mueller$^{1}$, Tobias Doernbach$^{1}$, Davide Chiarella$^{2}$ and Andreas Birk$^{1}$
\thanks{\textsuperscript{*}The research leading to the presented results has received funding from the German Federal Ministry of Education and Research within the project ``Vulnerability of Transportation Structures – Warning and Evacuation in Case of Major Inland Flooding'' (FloodEvac) and EU FP7 project ``Cognitive Autonomous Diving Buddy (CADDY)''}
\thanks{$^{1}$ Robotics Group, Computer Science \& Electrical Engineering, Jacobs University Bremen, Germany. \texttt{\{a.gomezchavez,\allowbreak chr.mueller,\allowbreak t.fromm,\allowbreak a.birk\}@jacobs-university.de}}%
\thanks{$^{2}$ Institute of Intelligent Systems for Automation—National Research Council of Italy, Genova, Italy. \texttt{davide.chiarella@cnr.it}}
}

\begin{document}

\maketitle
\thispagestyle{empty} 

\ieeefootline{Workshop on Human-Aiding Robotics \\ International Conference on Intelligent Robots and Systems 2018 (IROS), Madrid, Spain}


\begin{abstract}

We propose a robust gesture-based communication pipeline for divers to instruct an Autonomous Underwater Vehicle (AUV) to assist them in performing high-risk tasks and helping in case of emergency.
A gesture communication language (CADDIAN) is developed, based on consolidated and standardized diver gestures, including an alphabet, syntax and semantics, ensuring a logical consistency.
A hierarchical classification approach is introduced for hand gesture recognition based on stereo imagery and multi-descriptor aggregation to specifically cope with underwater image artifacts, e.g. light backscatter or color attenuation.
Once the classification task is finished, a syntax check is performed to filter out invalid command sequences sent by the diver or generated by errors in the classifier. 
Throughout this process, the diver receives constant feedback from an underwater tablet to acknowledge or abort the mission at any time. 
The objective is to prevent the AUV from executing unnecessary, infeasible or potentially harmful motions.
Experimental results under different environmental conditions in archaeological exploration and bridge inspection applications show that the system performs well in the field.

\end{abstract}

\section{INTRODUCTION}

Underwater environments pose a great number of technological challenges for robotics in the areas of navigation, communication, autonomy, manipulation and others. Although progress has been quickly made in the last years, the unstructured and dynamic nature of these environments makes human intervention indispensable for applications such as acquisition of relevant biological data, monitoring of marine areas, exploration of archaeological sites, inspection of damaged infrastructure and search and rescue operations.

\begin{figure}[H]
	\centering
	\captionsetup{justification=justified}
	\begin{subfigure}{\linewidth}
		\centering
		\includegraphics[width=.75\linewidth]{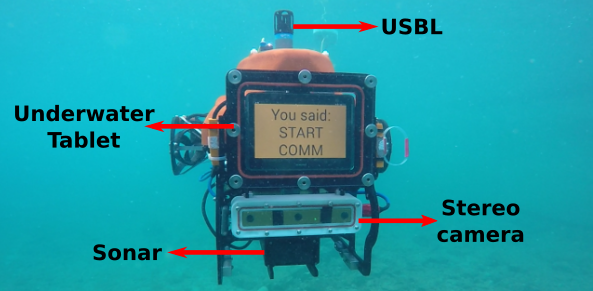}
	\end{subfigure}
	\begin{subfigure}{\linewidth}
		\centering
		\includegraphics[width=.75\linewidth]{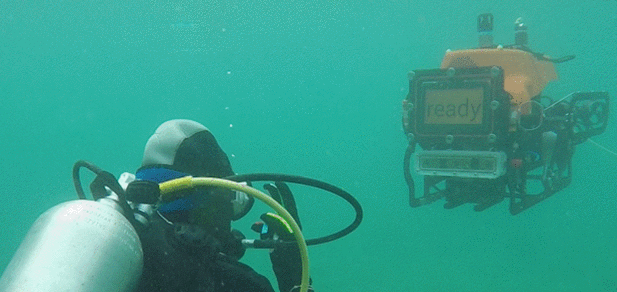}
	\end{subfigure}
	\caption{(Top) BUDDY-AUV equipped with sensors to monitor divers - Blueprint Subsea X150 USBL, ARIS 3000 Multibeam Sonar, BumbleBeeXB3 Stereo Camera, Underwater Tablet. (Bottom) Diver issuing gesture command to AUV~\cite{Miskovic2017_caddy-y3}.}
	\label{fig:buddy-auv}
\end{figure}

For this reason, the EU-funded Project CADDY (Cognitive Autonomous Diving Buddy) was developed with the aim of using robotic technology to improve the safety levels during underwater missions with divers in the loop. The main objective is the development of a pair of companion/buddy robots (Fig.~\ref{fig:buddy-auv}) -- an Autonomous Underwater Vehicle (AUV) and an Autonomous Surface Vehicle (ASV) -- to monitor and support human operations and activities during a dive. 

In underwater scenarios, there are unique sensor problems due to the attenuation of high-frequency electromagnetic waves: WiFi and radio signals become quickly unreliable at \SI{0.5}{m} depth, no GPS, and optical devices and camera imagery suffer artifacts due to light backscattering. Acoustic sensors are mostly used, although they offer limited bandwidth. Thus, the proposed solution is a communication framework letting the diver interact with the AUV using an extension of the common diver-gestures (CADDIAN language~\cite{Chiarella2015_CADDIAN}).
The diver gesture is captured using a stereo camera and process through a multi-descriptor classifier robust to underwater distortions. Then, when a complete message has been relayed, a syntax check is done to corroborate logical coherence and an approval request is issued to the diver via an underwater tablet (Fig.~\ref{fig:buddy-auv}). 

In this work, we show the application of the developed CADDY system into bridge inspection.
Particularly, we triggered the inspection procedures on~\cite{Mueller2017_BridgeInspection} as part of the FloodEvac BMBF research project$^*$. 
Bridges have an important role during disaster scenarios for evacuation and humanitarian supply; the remaining load capacity of bridges after a disaster is a crucial factor to analyze. 
In order to obtain data about structural components and the situation of foundations, the current state-of-the-art procedure relies on diver operations. However, this is an undertaking high risk in flooded areas, and commonly divers can only perform haptic surveys of the damaged areas due to poor visibility.

\section{SYSTEM OVERVIEW}

The two main objectives of the system are: generation of AUV missions consisting of multiple commands derived from hand gestures and high fault-tolerance. 
The latter goal is of great importance in field underwater applications since recovery or maintenance of the equipment during a mission is not an easy task; and when there is a close human-robot interaction, the safety of the user is top priority.
 
For these reasons, the system architecture presented in Fig.~\ref{fig:general_framework_diagram} consists of several modules that ensure that the issued diver commands are logically consistent and scalable (stack up to form more complex missions), and that useful feedback is produced to make corrections on the fly if necessary. At the algorithmic level, these modules ensure that the recognition of the gestures are robust against sudden movements, light backscattering and poor contrast i.e., image distortions; and that the commanded missions are carried out by the AUV successfully by following the correct motion primitives. We describe the functionality of each of these modules as follows: 

\begin{figure}[t]
	\centering
	\includegraphics[width=0.99\linewidth]{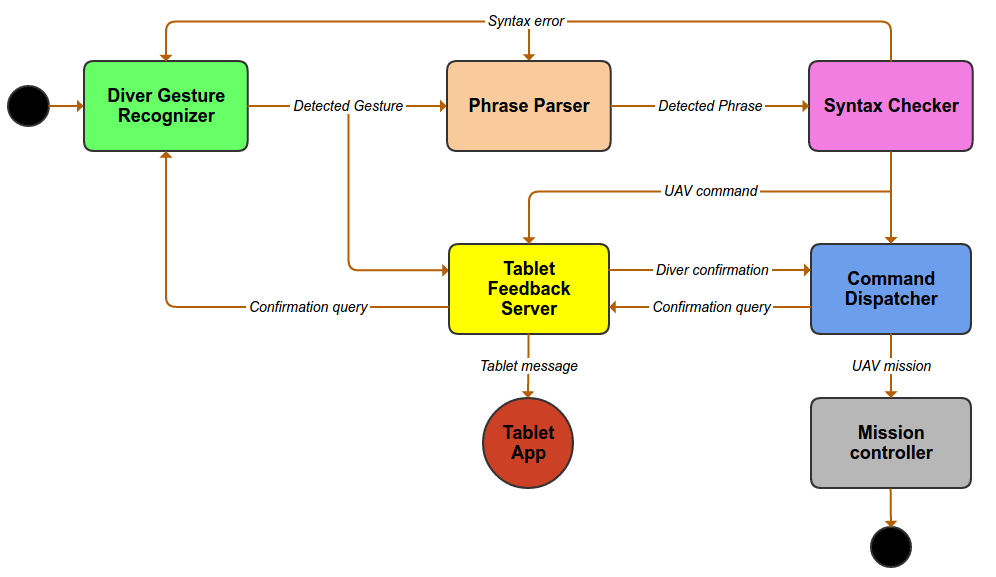}
	\caption{System architecture of AUV mission generator based on diver hand gestures.} 
	\label{fig:general_framework_diagram}
\end{figure}

\begin{itemize}
	\item \textit{Diver Gesture Recognizer} contains all the image processing and classification algorithms to correctly detect the diver's hands and determine if they correspond to a known command for the AUV. The output is a label indicating the command's name. 
	
	\item \textit{Phrase Parser} receives the previous labels and parse them according to some delimiters i.e., special gestures that group basic commands into sequences named \textit{phrases} (Fig.~\ref{fig:parser}).
	
	\item \textit{Syntax Checker} evaluates the previous \textit{phrases} according to the CADDIAN language syntax rules (Section~\ref{sec:caddian_language}). In this way, only \textit{complete} and \textit{logically consistent} commands are saved in memory for future execution. For example the command ``Do a Map and Go Down'' will not be accepted since both actions need parameters to be executed, i.e. ``Do a Map of 10 by 12 meters and Go Down 1 meter.''
	
	\item \textit{Command Dispatcher} receives the validated AUV commands from the \textit{Syntax Checker} and stores them until the diver gives a final approval command. Then, it feeds these commands sequentially to the \textit{Mission Controller} as each of them is successfully completed. 
	
	\item \textit{Mission Controller} maps the received AUV commands into low-level tasks and monitors the activation and execution of the necessary action primitives to complete them.
	
	\item \textit{Tablet Feedback Server} prepares the message that is going to be displayed in the Tablet according the input from the previous the modules.
\end{itemize}

\section{Diver Gesture Recognizer}

\begin{figure}[t]
	\centering
	\includegraphics[width=0.99\linewidth]{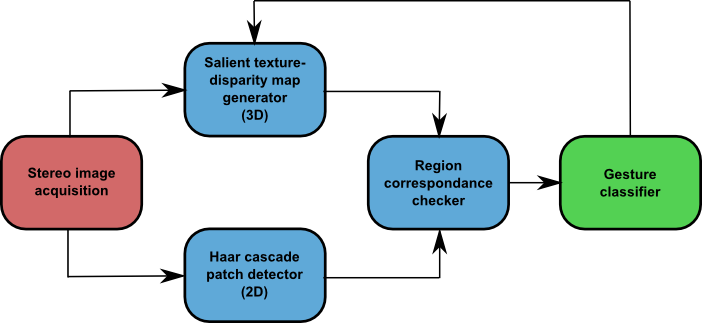}
	\caption{General framework diagram for hand gesture recognition. The different stages of the process are color coded: (Red) Image acquisition, (Blue) Hand detection, (Green) Gesture classification.} 
	\label{fig:hand_gesture_recognition_framework}
\end{figure}

Hand gesture recognition applications have achieved high precision and performance through the use of RGB-D cameras that generate dense point clouds from which 3D features can be computed and reasoned about. However, this technology is not best suited for underwater applications due to light backscatter in water~\cite{Sun2013}. The use of stereo cameras allows to retrieve more accurate 3D information, but the generated point clouds are sparse and the current state of the art gesture recognition methods do not perform well on these data. 

Thus, a hybrid-method using 3D and 2D information was implemented as it was found to yield the best results for object detection and recognition tasks in the highly variant underwater imagery. Figure~\ref{fig:hand_gesture_recognition_framework} shows the diagram of the overall hand gesture recognition framework; there are two main stages: \textit{hand detection} and \textit{gesture classification}. 

For \emph{detection}, we take the hybrid approach combining disparity maps and cascade classifiers to ensure robustness against motion and light attenuation. Segmentation of disparity maps based on distance and density offers reliable hand detection, however, it fails in the presence of bubbles and texture-prominent areas. Hence, we use 2D cascade classifiers to filter out these false positive regions. 

\begin{figure}[t]
	\centering
	\includegraphics[width=0.98\linewidth]{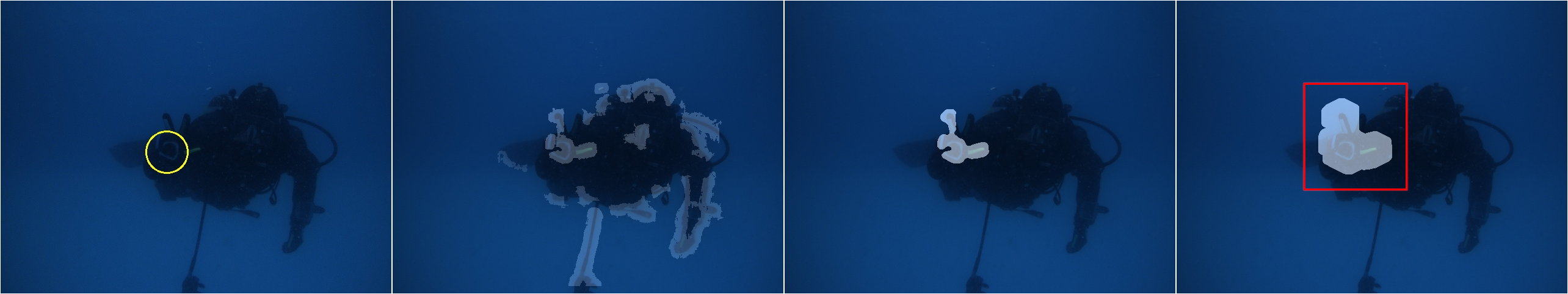}
	\caption{(First column) Detected hand candidate regions by cascade classifier (Second column) Disparity points generated from the stereo pairs (Third column) Disparity points after median filter (Fourth column) Salient texture regions found after erode-dilate filters are applied. } 
	\label{fig:gesture detection}
\end{figure}

\begin{figure}[t]
	\centering
	\includegraphics[width=0.99\linewidth]{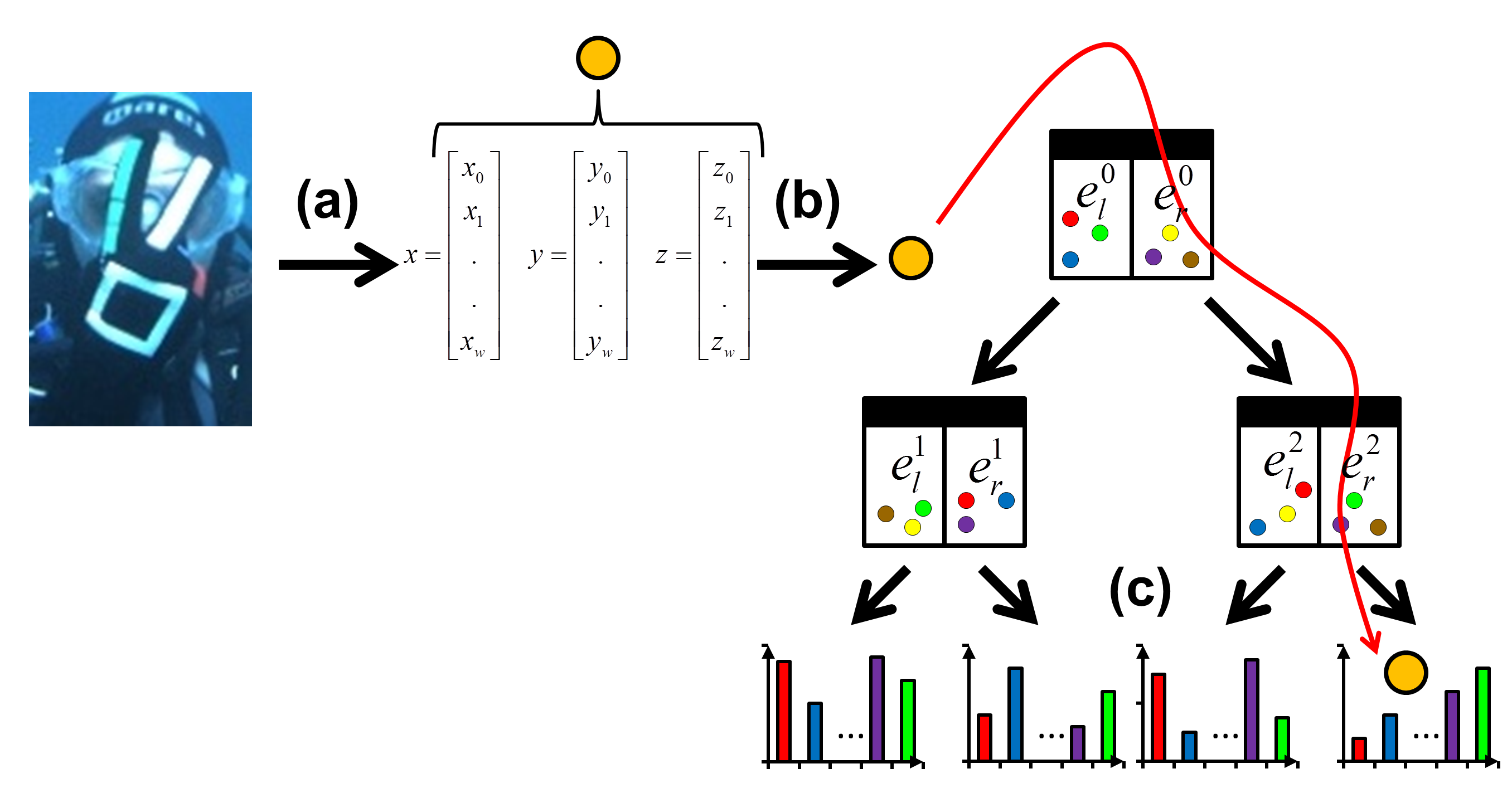}
	\caption{Classification of an image by a Multi-Descriptor NCM tree (MD-NCM)~\cite{Gomez2015_DiverDetection}; (a) feature centroid is extracted represented by a color dot and (b) it traverses the path through the decision tree until (c) classified.} 
	\label{fig:ncmf_diagram}
\end{figure}

Once the system has detected possible locations of the diver's hands within the image (Fig.~\ref{fig:gesture detection}), all of these candidate patches pass through a final classifier: Multi-Descriptor Nearest Class Mean Forests (NCMFs), which was first introduced for diver localization in our previous work in~\cite{Gomez2015_DiverDetection}. This classifier filters out all the false positives generated by the previous modules and maps the true positives (hands) to a specific gesture within the CADDIAN language. The main purpose of this proposed variant of Random Forest is to aggregate multiple descriptors that encode different representations of the objects of interest, each of them robust to different types of image distortion (Fig.~\ref{fig:ncmf_diagram}).

It is important to note that this hybrid method was used and developed throughout the duration of the CADDY and FloodEvac projects due to its robustness given small sample datasets from underwater environments. After several field trials and data collection efforts, a detailed dataset with stereo images from the divers communicating with the AUV in different environmental conditions has been assembled (\href{http://caddy-underwater-datasets.ge.issia.cnr.it/}{CADDY-Dataset}~\cite{GomezChavez2018_CADDY-Dataset}). Based on this, we hope to further investigate underwater image artifacts and enable data-driven methods.

\section{CADDIAN LANGUAGE}
\label{sec:caddian_language}

\begin{figure}[t]
	\centering
	\begin{subfigure}{0.24\linewidth}
		\includegraphics[width=\linewidth]{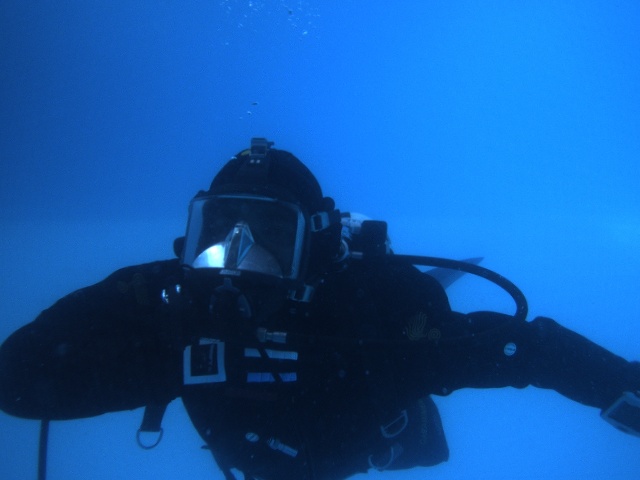}
		\caption{Out of air}
	\end{subfigure}
	\begin{subfigure}{0.24\linewidth}
		\includegraphics[width=\linewidth]{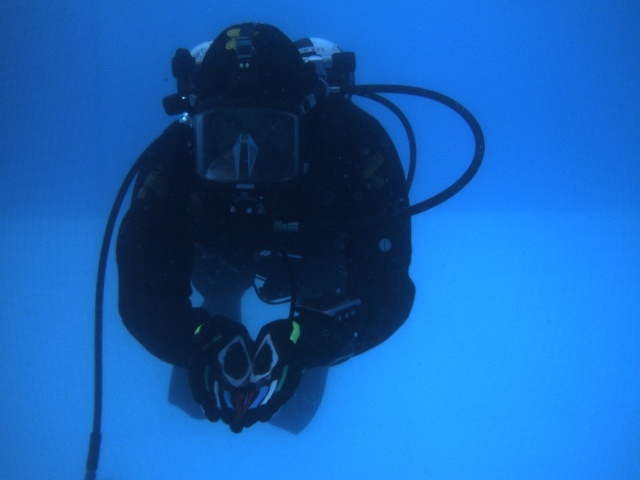}
		\caption{Boat}
	\end{subfigure}
	\begin{subfigure}{0.24\linewidth}
		\includegraphics[width=\linewidth]{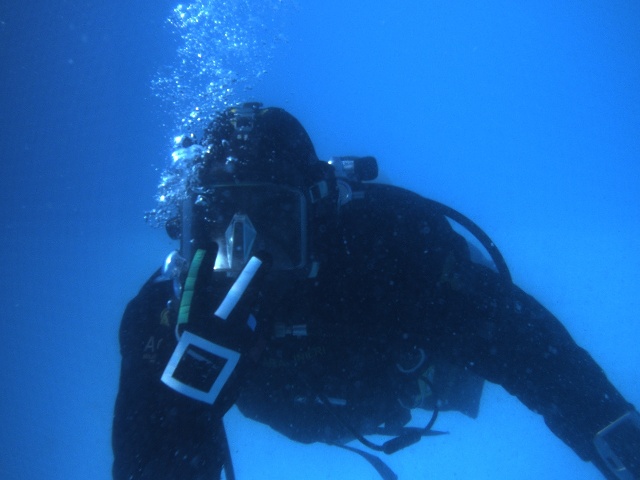}
		\caption{Start comm}
	\end{subfigure}
	\begin{subfigure}{0.24\linewidth}
		\includegraphics[width=\linewidth]{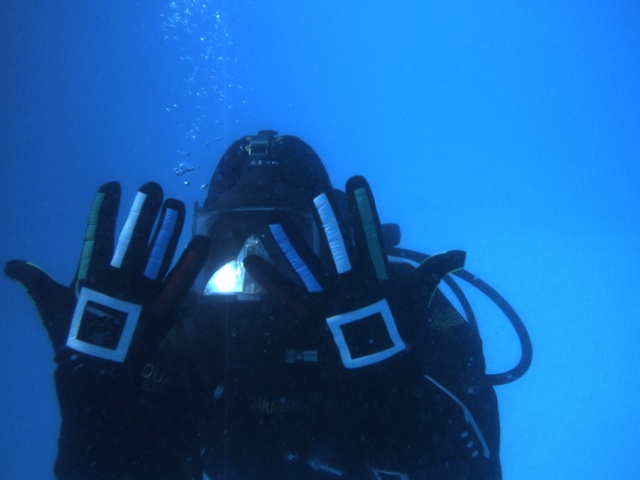}
		\caption{Do a mosaic}
	\end{subfigure}
	
	\caption{Examples of CADDIAN gestures. (a) and (b) are extracted from already established gestures. (c) and (d) were defined to be easily related to the actions they represent.}
	\label{fig:examples_caddian}
\end{figure}

As described before, the underwater environment is an hazardous scenario, where communication between peers is crucial for the success and safety of the dive.
The developed CADDIAN gestures~\cite{Chiarella2015_CADDIAN} were chosen and/or defined from those common to divers to be as intuitive as possible, reduce learning time and transmit messages efficiently. However, diver's hand signals vary from region to region and between organizations, hence the most common ones were selected from ~\cite{CMAS}~\cite{NEADC}~\cite{scubadivingfanclub}. Figure~\ref{fig:examples_caddian} shows examples of CADDIAN gestures, the first two gestures are used in recreational diving while the last two were defined to be easily related to the action they are representing: the \textit{Start Communication} gesture consists of pointing the index and middle finger to the mask, which symbolizes \textit{``Look at me!''}; likewise, the \textit{Do a mosaic/map} gesture mimics a person holding a map in front of him and reading it. 

\subsection{SYNTAX CHECKER}

We described the methodology and algorithms used to detect and classify single gestures. However, to exploit the full potential of the application, the system must be able to understand complex commands - sequences of gestures - which can be also aggregated to form missions composed of several tasks. To achieve this and allow the diver to understand the status and progress of a mission, a communication protocol was established.

To allow synchronization between the diver and the AUV feedback messages, the CADDIAN syntax defines boundaries in order to ensure correct interpretation of commands and missions. These boundaries are associated with the gestures ``START\_COMMUNICATION'' or $A$ and ``END\_COMMUNICATION'' or $\forall$; thus, commands are sequences of individual gestures delimited by $(A,A)$ which commonly represent a single task, e.g. ``Take a photo at 3 meters altitude'', and missions are delimited by $(A,\forall)$ which consist of aggregated commands, e.g. ``Take a photo here, go to the boat and carry the equipment back here". Of course, a mission composed of a single command is valid.

\begin{figure}[t]
	\centering
	\includegraphics[width=1\columnwidth,keepaspectratio]{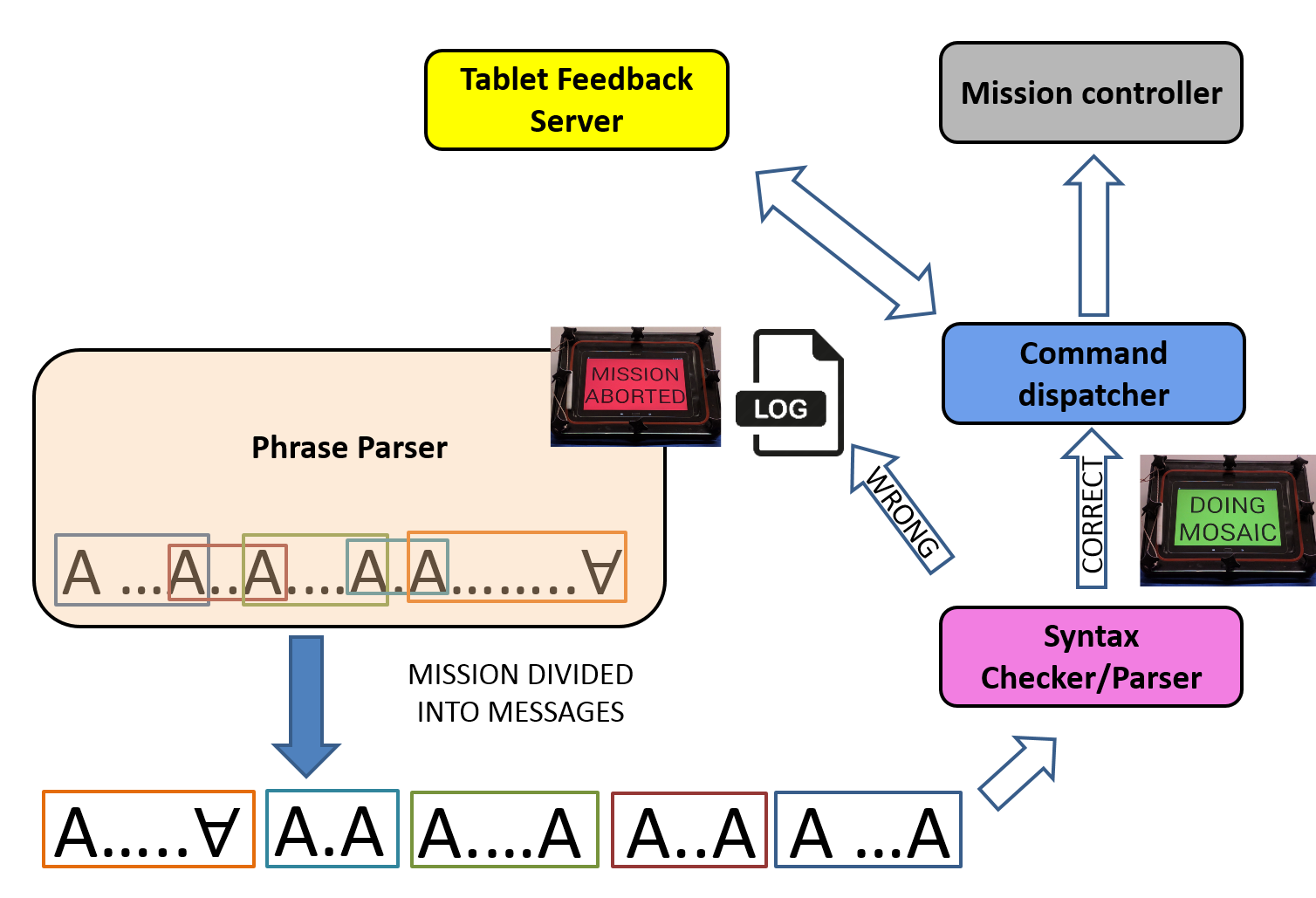}
	\caption{Mission segmentation into single messages which are processed and validated by the Syntax Checker.} \label{fig:parser}
\end{figure}

The gesture sequence interpretation is implemented through the \textit{Phrase Parser} module introduced in Fig.~\ref{fig:general_framework_diagram}. It constantly saves the detected gestures until it detects among them one of the mentioned delimiter pairs $(A,A)$, $(A,\forall)$; then it sends these gestures - commands - to the \textit{Syntax Checker} for validation. If the command is syntactically correct, it passes to the \textit{Command Dispatcher} module where it is saved until a complete mission is received i.e. an ``END\_COMMUNICATION'' gesture issued by the diver. Finally, these commands are sequentially passed to the \textit{Mission Controller} for execution. On the other hand, if the command is not valid, the error is logged and a warning is issued to the diver through the underwater tablet or a system of lights, Fig.~\ref{fig:parser} depicts this process.

\section{BRIDGE INSPECTION APPLICATION}

\begin{figure}[t]
	\centering
	\captionsetup{justification=justified}
	\begin{subfigure}{\linewidth}
		\centering
		\includegraphics[width=.99\linewidth]{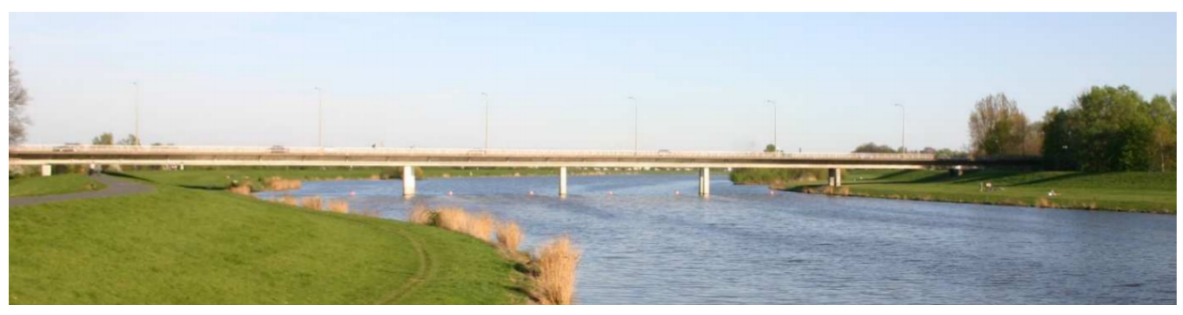}
	\end{subfigure}
	\begin{subfigure}{\linewidth}
		\centering
		\includegraphics[width=.99\linewidth]{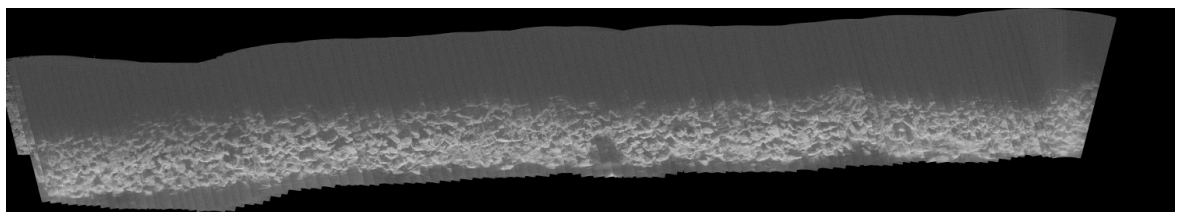}
	\end{subfigure}
	\caption{(Top) Karl Carstens Bridge and (Bottom) Registration sample of lakeshore section beneath the bridge using \emph{Fourier-Mellin Invariant} (FMI) method~\cite{Heiko2009}.}
	\label{fig:bridge}
\end{figure}

In flood scenarios, disaster management, e.g. led by government authorities, demands accurate assessment of the vulnerability of infrastructures like bridges in order to plan safe evacuation.
Given a bridge inspection request, the vehicle in Fig.~\ref{fig:buddy-auv} its deployed along with a diver. In our test deployment scenario the Karl Carstens bridge in Bremen, Germany (Fig.~\ref{fig:bridge}) has been selected.

In this case the sonar-based camera ARIS (Adaptive-Resolution Imaging Sonar) is used to map the foundations of the bridge due to poor visibility. If the diver is unsure whether it is safe to approach the bridge, he can issue a command \emph{Do a map of X and Y dimensions} under the bridge. Then, after visualizing the images via the underwater tablet or in an on-shore center (where another expert is located), he can decide to approach the bridge for a more in detailed inspection.
Frequently, analysts are in the look out for \emph{scouring} (soil erosion) and \emph{log-jams} (piled-up wood). In our situation, since no debris was found, the diver decided he could safely analyze the structure up-close (Fig.~\ref{fig:bridge}).

Based on this first-hand assessment and the gathering of other measurements such as \emph{flow velocity} through a discharge sensor to analyze hydrodynamic forces; a final veredict about the state and usability of the bridge can be done. 

\section{CONCLUSIONS}

Despite the high risk that involves disaster scenarios in underwater environments, full autonomy has not been achieved by ROVs and/or UAVs; thus, human (diver) experts are needed to perform high precision tasks or assess the stability of submerged structures. We think the major challenge consists in the development of a communication protocol between robot and diver to actively interact and cooperate and, above all, ensure the safety of the diver. Based on this and previous work, and on the bridge inspection scenario presented, we believe that symbiosis between human experts and robots is the trigger to spread the usage of this technology; particularly in these applications, where not all possible cases can be predicted.  

\addtolength{\textheight}{-12cm}   


\bibliographystyle{IEEEtran}
\bibliography{bibliography}

\end{document}